\documentclass[iicol]{sn-jnl}
\usepackage{graphicx}%
\usepackage{multirow}%
\usepackage{amsmath,amssymb,amsfonts}%
\usepackage{amsthm}%
\usepackage{mathrsfs}%
\usepackage[title]{appendix}%
\usepackage{xcolor}%
\usepackage{textcomp}%
\usepackage{manyfoot}%
\usepackage{booktabs}%
\usepackage{algorithm}%
\usepackage{algorithmicx}%
\usepackage{algpseudocode}%
\usepackage{listings}%
\theoremstyle{thmstyleone}%
\theoremstyle{thmstyletwo}%

\theoremstyle{thmstylethree}%

\begin{document}

\title[Article Title]{3DProxyImg: Controllable 3D-Aware Animation Synthesis from Single Image via 2D-3D Aligned Proxy Embedding}


\author[1]{\fnm{Yupeng} \sur{Zhu}}\email{zhuyupeng@sjtu.edu.cn}
\equalcont{These authors contributed equally to this work.}

\author[1]{\fnm{Xiongzhen} \sur{Zhang}}\email{1492505833zhang@gmail.com}
\equalcont{These authors contributed equally to this work.}

\author[1]{\fnm{Ye} \sur{Chen}}\email{chenye123@sjtu.edu.cn}
\equalcont{These authors contributed equally to this work.}

\author*[1]{\fnm{Bingbing} \sur{Ni}}\email{nibingbing@sjtu.edu.cn}

\affil[1]{\orgdiv{School of Electronic Information and Electrical Engineering,  Shanghai Jiao Tong University, Shanghai 200240, China}}

% \affil[2]{\orgdiv{Department}, \orgname{Organization}, \orgaddress{\street{Street}, \city{City}, \postcode{10587}, \state{State}, \country{Country}}}

% \affil[3]{\orgdiv{Department}, \orgname{Organization}, \orgaddress{\street{Street}, \city{City}, \postcode{610101}, \state{State}, \country{Country}}}


\abstract{3D animation is central to modern visual media, yet traditional production pipelines remain labor-intensive, expertise-demanding, and computationally expensive. Recent AIGC-based approaches partially automate asset creation and rigging, but either inherit the heavy costs of full 3D pipelines or rely on video synthesis paradigms that sacrifice 3D controllability and interactivity. We focus on single-image 3D animation generation and argue that its progress is fundamentally constrained by a trade-off between rendering quality and 3D control.
To address this limitation, we propose a lightweight 3D animation framework that decouples geometric control from appearance synthesis. The core idea is a 2D–3D aligned proxy representation that uses a coarse 3D estimate as a structural carrier, while delegating high-fidelity appearance and view synthesis to learned image-space generative priors. This proxy formulation enables 3D-aware motion control and interaction comparable to classical pipelines, without requiring accurate geometry or expensive optimization, and naturally extends to coherent background animation. Extensive experiments demonstrate that our method achieves efficient animation generation on low-power platforms, and outperforms video-based 3D animation generation in identity preservation, geometric and textural consistency, and the level of precise, interactive control it offers to users.}

\keywords{3D Animation, Proxy Embedding, Single-image to 3D}



\maketitle

\section{Introduction}\label{sec:intro}
Image animation~\cite{siarohin2019first,yin2023nerfinvertor,wu2025aniclipart,hu2024animate,xu2024magicanimate,guo2023animatediff,tan2024animate} plays a pivotal role in contemporary visual creation and interactive experiences, with extensive applications across entertainment sectors such as film, gaming, and virtual reality. Unlike traditional 2D animation~\cite{siarohin2019first,yin2023nerfinvertor,wu2025aniclipart}, 3D-aware animation is characterized by its ability to create a sense of depth and volume, offering a more lifelike and immersive representation of objects and environments. This three-dimensional quality enhances spatial perception, enabling dynamic interactions and more realistic visual storytelling.

Traditional 3D animation typically follows a fixed pipeline that includes 3D modeling, texturing and material setup, rigging and skinning, keyframe or motion-capture–based animation, and finally lighting and rendering~\cite{au2008skeleton,xu2020rignet,baran2007automatic,song2025magicarticulate,song2025puppeteer}. This paradigm offers high controllability and rich interactivity, since artists can precisely manipulate every stage of the process. However, it also makes 3D content creation extremely heavy: the workflow is labor-intensive, time-consuming, and requires specialized expertise and substantial computational resources, which severely limits scalability and rapid prototyping.

In the AIGC era, many components of this traditional pipeline are being partially replaced or accelerated by AI algorithms. For example, 3D generative models can already synthesize high-quality assets efficiently~\cite{wu2024direct3d,zhang2024clay,xiang2025structured}, and recent learning-based methods have made progress on automatic rigging and skinning~\cite{song2025magicarticulate,song2025puppeteer}. Nevertheless, due to the intrinsic complexity of the 3D animation pipeline, current AI techniques still cannot cover the whole production process. For instance, Puppeteer can automatically rig and skin arbitrary meshes, but it still relies on computationally expensive, long-horizon optimization over video priors to drive the skeleton, which not only demands heavy computation but also sacrifices the controllability and interactivity that are central to professional 3D animation. Therefore, this work is motivated to develop an AI-assisted 3D generation paradigm that is lightweight, highly controllable, and strongly interactive. Similarly, 3D animation generation based on video synthesis algorithms~\cite{hu2024animate,tan2024animate} loses interactivity during the animation production process, making it difficult to be widely applied.

Building on this observation, we argue that progress in single-image 3D animation generation is fundamentally constrained by a trade-off between rendering quality and 3D controllability. Classical 3D animation pipelines demand accurate geometry and skinning, since coarse models rapidly degrade visual fidelity, whereas video-based animation paradigms can produce high-quality frames but offer limited 3D control and interactivity. To bridge this gap, we introduce a lightweight animation pipeline based on a 2D–3D aligned proxy embedding, which enables high-quality, interactive 3D animation from only a coarse 3D estimate. Given an input image, we first recover a 3D shape whose position and scale are aligned with the target object by combining monocular depth estimation~\cite{wang2025vggt,yang2024depthv2} with image-to-3D generative modeling~\cite{zhao2025hunyuan3d}. Rather than directly projecting this imperfect geometry—with inaccurate shape and texture—into the image plane, we follow proxy-based implicit rendering~\cite{chen2025easy,Chen_2024_CVPR,mildenhall2020nerfrepresentingscenesneural} to transfer 3D information into the image domain. Concretely, we sample sparse vertices (proxy nodes) on the 3D surface and attach distributed implicit texture features to them. After projecting these nodes into the 2D image, the sparse features locally modulate the rendering around each node, substantially reducing artifacts caused by geometric and textural errors. To recover textures that are invisible in the input view (e.g., on the back of the object) while preserving consistency with the observed image, we apply random affine transformations to the 3D object, re-render the resulting views into the image space, and constrain the synthesis with generative priors, resulting in high-fidelity, multi-view consistent textures.

It is worth noting that although the proxy nodes are sparse, they are sufficient to represent the 3D geometry of the target object. As a result, we can directly apply skinning and drive them to generate animation sequences with a 3D feel. Thanks to the texture representation based on distributed proxy embedding and implicit rendering techniques, our approach achieves excellent texture accuracy and consistency during the target object-driven process. Furthermore, for the background components outside the driving target, we propose a proxy representation propagation method for background completion, which enables efficient and realistic animation rendering results. Extensive experiments demonstrate that, compared to traditional 3D animation pipelines, our method achieves efficient animation generation on low-power platforms. In contrast to video-based 3D animation generation methods, our approach ensures high consistency in target ID, geometry, and texture, while providing users with a highly accurate and controllable interactive experience.

% Building upon the proposed 2D–3D aligned proxy geometry and texture representations, our method enables realistic and high-fidelity 3D-aware image editing through explicit manipulations of the underlying 3D shape using simple operations (\emph{e.g.}, rotation and rescaling). Extensive experiments and visualizations validate the effectiveness of our approach, showing that it achieves superior controllability and multi-view consistency compared to state-of-the-art methods, especially when applied to repeated editing on the same image. Editing results as shown in Fig.~\ref{fig: example}.

\section{Related Works}\label{sec:rw}
\noindent\textbf{Neural Image Representation.} Implicit neural representations (INRs) have become a foundational tool for high-fidelity 3D reconstruction and view synthesis~\cite{atzmon2020sal,michalkiewicz2019implicit,gropp2020implicit,jiang2020local,chabra2020deep,sitzmann2019scene}. NeRF~\cite{mildenhall2021nerf} and its variants such as Instant-NGP~\cite{muller2022instant} represent scenes as continuous functions, mapping spatial coordinates and viewing directions to color and density values. These methods demonstrate impressive novel view synthesis capabilities but typically require dense multi-view supervision and extensive computation. To alleviate these constraints, recent approaches explore generative priors or learned scene priors for single-image reconstruction~\cite{ulyanov2018deep, sitzmann2020implicit, muller2022instant,chen2021learning, chen2024image, chen2024towards}, enabling compact image representation and convenient editing. Additionally, hybrid techniques like proxy-based implicit rendering attach learnable texture or appearance features to sparse anchors (e.g., surface points or mesh vertices), which are projected into the image domain to guide rendering. This allows decoupling appearance modeling from precise geometry, making it more robust to imperfect shape estimates~\cite{chen2024image, chen2024towards}. Our work follows this direction and introduces a lightweight, distributed proxy embedding that facilitates high-quality, 3D-consistent rendering while maintaining low computational overhead.

\noindent\textbf{Image Animation.} Image animation methods span 2D warping~\cite{siarohin2019first,yin2023nerfinvertor,wu2025aniclipart} and generative video synthesis~\cite{hu2024animate,xu2024magicanimate,guo2023animatediff,tan2024animate}. 2D-based approaches animate static images by estimating motion through keypoints or optical flow. Notable works like First Order Motion Model (FOMM)~\cite{siarohin2019first} learn unsupervised keypoints and local transformations to drive deformation, while later variants~\cite{zhao2022thin} introduce improved motion fields and occlusion handling. These methods are efficient and general but often lack temporal stability and fail under large pose or viewpoint changes due to their lack of structural understanding. Recently, image-to-video generation using diffusion models (e.g., AnimateDiff~\cite{guo2023animatediff}, Animate Anyone~\cite{hu2024animate}) has emerged as a powerful direction, generating realistic motion sequences from static images and motion prompts. While these models produce visually appealing results, they often trade off explicit control and 3D consistency for perceptual quality. In contrast, our method combines sparse geometry and proxy-based implicit rendering to achieve accurate, interactive, and 3D-consistent animation with low computational cost.

\section{Methodology}\label{sec:method}
\begin{figure*}[t]
    \centering
    \includegraphics[width=1.0\linewidth]{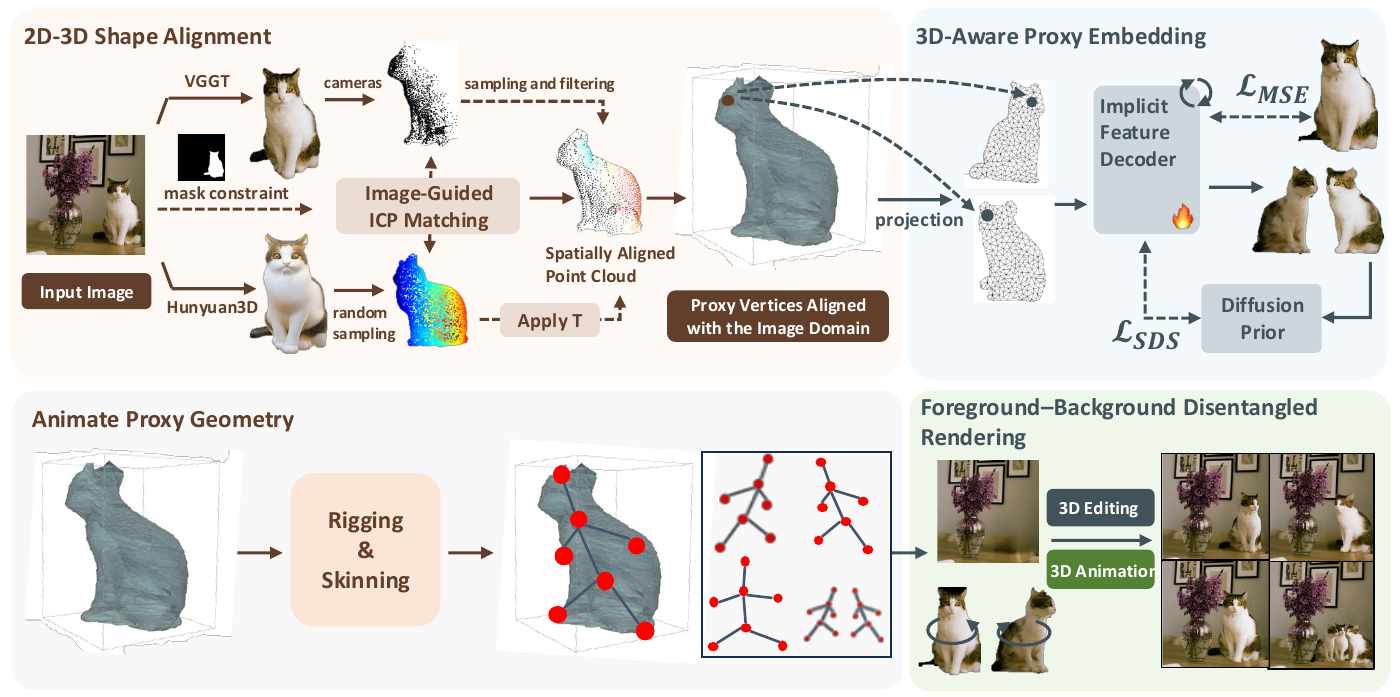}
    \caption{The pipeline of the framework. Given an image, our framework first integrates 3D reconstruction and generative models to obtain a geometrically coherent representation with spatially aligned 2D-3D shapes. Subsequently, through implicit neural rendering and SDS optimization, we derive a 3D asset exhibiting high-fidelity texture and multi-view consistency, which is then utilized to guide the 3D-aware editing and animation process of 2D image.}
    \label{fig: pipeline}
\end{figure*}

This paper presents a novel framework for generating high-fidelity and multi-view consistent 3D animation from a single input image. An overview of our pipeline is shown in Fig.~\ref{fig: pipeline}. Our method first tackles the misalignment between coarse 3D representations and the original 2D image domain by synergistically integrating monocular depth estimation and generative 3D models. We then construct a hybrid proxy representation that leverages an implicit neural feature field for high-fidelity texture rendering. Finally, we utilize the powerful texture and geometric priors of a pretrained 2D diffusion model~\cite{esser2024scaling} to enhance detail fidelity and enforce multi-view consistency via Score Distillation Sampling (SDS)~\cite{poole2022dreamfusiontextto3dusing2d}. The reconstructed proxy geometric representation is then rigged, skinned, and driven, while rendering with the distributed proxy-embedding representation enables the synthesis of highly controllable, high-fidelity 3D animations at low computational cost.

\subsection{2D-3D Shape Alignment}
\label{subsec: 1}
Performing 3D-form editing (e.g., 3D rotation, dragging, and scaling) of an object in a given image remains a significant challenge. The core challenge of this task lies not only in recovering complete and accurate 3D geometry from the limited 2D observation but also in accurately projecting the acquired 3D asset back into the 2D image space to explicitly guide the editing of image content with high controllability.

Current state-of-the-art methods~\cite{li2025sparc3d,voleti2024sv3d,zhao2025hunyuan3d} for generating 3D assets from a single image commonly exhibit the following limitations: (1) The texture representations are often coarse, tending to reconstruct smooth low-frequency surfaces, which makes it difficult to faithfully recover high-frequency textural details of complex materials such as fur, wood, and leather; (2) The overall scale and local structural proportions (for example, between the limbs and the body) of the generated geometry often deviate from the actual appearance of the object in the known 2D observation, and pose estimation may also be inaccurate; (3) 3D assets generated by different models usually lack uniformity in coordinate systems and scales, and typically do not provide projection parameters (such as camera parameters) directly associated with the input image. Consequently, even when a 3D representation of the target object is obtained, it remains challenging to accurately project it into the original 2D image space, thereby hindering effective support for image editing and reconstruction at the 3D level.

To align 3D assets with the 2D image domain, we decompose the 3D reconstruction task into two distinct stages: coordinate-aligned reconstruction for known viewpoints and coarse reconstruction with alignment for unknown viewpoints. For the single viewpoint corresponding to the input image $I$, we use the state-of-the-art 3D reconstruction model VGGT~\cite{wang2025vggt} to obtain a precise, pixel-aligned 3D point cloud, denoted as $\mathbf{P}_{vggt}\in\mathbb{R}^{N \times 3}$, while it lacks complete information on the occluded rear portions. Then we leverage the powerful single-view 3D generation model, Hunyuan3D~\cite{zhao2025hunyuan3d}, to acquire a full mesh representation and uniformly sample a point cloud from the mesh surface, denoted as $\mathbf{P}_{hy}\in\mathbb{R}^{N \times 3}$. For the aforementioned misalignment issue, we designed a points matching pipeline as follows:
\begin{equation}
    \mathbf{T} = ICP(\mathbf{P}_{vggt},\mathbf{P}_{hy}),
    \label{equ: 1}
\end{equation}
\begin{equation}
    \mathbf{P^\prime}_{hy} = \mathbf{T}(\mathbf{P}_{hy}),
    \label{equ: 2}
\end{equation}
where ICP is the iterative closest point algorithm~\cite{besl1992method}, $\mathbf{T}$ is the transformation matrix, and $\mathbf{P^\prime}_{hy}$ is points after applying $\mathbf{T}$. We fuse $\mathbf{P}_{vggt}$ and $\mathbf{P^\prime}_{hy}$, and introduce a Laplacian constraint to eliminate discontinuities.

To address the inaccuracies in the ICP solution, we project the fused points back into the 2D image plane using camera parameters in VGGT. The error between the original object mask $M_{obj}$ and the projected mask $M_{proj}$ is then calculated to further optimize $\mathbf{T}$, ultimately achieving an optimal alignment $\mathbf{P}_{aligned}$. $\mathbf{P}_{aligned}$ can be accurately projected back into the 2D image space, thereby establishing a precise geometric foundation for subsequent stages.
\begin{equation}
    \mathbf{T}_{opt} = \operatorname*{argmin}_{T} \|M_{obj} - M_{proj}\|,
    \label{equ: 3}
\end{equation}
\begin{equation}
    \mathbf{P}_{aligned} = fuse(\mathbf{P}_{vggt},\mathbf{T}_{opt}(\mathbf{P}_{hy})).
    \label{equ: 4}
\end{equation}

\subsection{3D-Aware Proxy Embedding}
\label{subsec: 2}
% As shown in Equ.~\ref{equ: 1},
% \begin{equation}
%     f(A) = B,
%     \label{equ: 1}
% \end{equation}
% where $f$ is the projection fuction.
To enable high-fidelity rendering of the reconstructed 3D point cloud, inspired by Chen et al.~\cite{chen2025easy}, we introduce a hybrid 2D–3D neural representation that explicitly models geometric structure while implicitly encoding texture information.

Specifically, we first downsample the dense point cloud $\mathbf{P}_{aligned}$ into a set of sparse 3D proxy vertices $\mathbf{V}$ and establish their 3D connectivity as triangles. For each vertex $v_i$, we initialize an optimizable texture feature vector $f_i\in\mathbb{R}^{d}$. To effectively represent high-frequency textural details, inspired by NeRF~\cite{mildenhall2020nerfrepresentingscenesneural}, we apply positional encoding to the feature vectors:
\begin{align}
\small
\gamma(f) &=( \sin(2^{0}\pi f), \cos(2^{0}\pi f), \cdots, \\
&\sin(2^{L-1}\pi f), \cos(2^{L-1}\pi f)),
    \label{equ:5}
\end{align}
where $\gamma(\cdot)$ is the encoding function. %We set $L=10$ in experiments.
The triangular faces are sorted by their depth from the given viewpoint $\beta$, after which the observable points from $\mathbf{V}$ are projected onto the 2D plane as $V_{\beta}$ with camera parameters provided by VGGT:
\begin{equation}
    V_{\beta} = Proj(\mathbf{V},\beta).
    \label{equ:6}
\end{equation}
The region defined by $V_{\beta}$ constitutes the target area for rendering the object.
For each pixel $p_i$ within the object region, the corresponding triangle is first determined, and then the barycentric coordinates of the pixels within the triangle are utilized to interpolate the feature vectors of the three vertices: 
\begin{equation}
    f_{p_i} =\lambda_af_a+\lambda_bf_b+\lambda_cf_c,
    \label{equ:7}
\end{equation}
where $f_{p_i}$ is the feature of the pixel $p_i$, $(f_a,f_b,f_c)$ are features on the vertices of a triangle and $(\lambda_a,\lambda_b,\lambda_c)$ are barycentric coordinates. The feature is then decoded into color values:
\begin{equation}
    (r,g,b) = D_{\theta}(f_{p_i}),
    \label{equ:8}
\end{equation}
where $D(\cdot)$ is a decoder composed of MLPs, which decodes the implicit features into color values, and $\theta$ represents the trainable parameters of the decoder.

Our experiment has proved that this representation and rendering method effectively combines 3D geometric information with the rich texture details of 2D observations.

\begin{figure*}[t]
    \centering
    \includegraphics[width=1.0\linewidth]{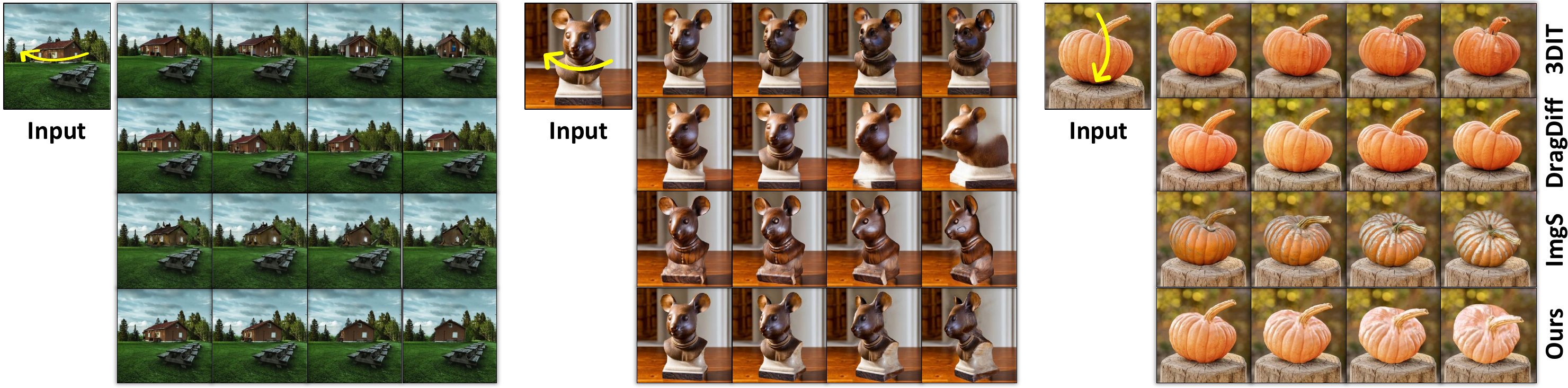}
    \vspace{-0.2cm}
    \caption{Visual comparison of different methods. Please zoom in for details.}
    \label{fig: quality}
    \vspace{-0.2cm}
\end{figure*}

After embedding texture features into 3D proxy nodes, we optimize the implicit features and the decoder to recover textural details of the 3D geometry. Two key aspects must be considered: (1) Ensuring the fidelity of the rendered object to the original image under known viewpoints; (2) Guaranteeing high-quality texture generation and multi-view consistency for the object under novel viewpoints. In view of these, we introduce two loss functions. Under known viewpoints $\beta_{ref}$, we calculate the MSE loss between the rendered image $I_{\beta_{ref}}$ and the input image $I$:
\begin{equation}
    \mathcal{L}_{\text{MSE}} = \|I_{\beta_{ref}} - I\|_2^2.
    \label{equ:9}
\end{equation}

For novel viewpoints, we introduce SDS~\cite{poole2022dreamfusiontextto3dusing2d} for texture generation. In each iteration, a random camera pose is sampled for rendering, and the prior from a 2D diffusion model is leveraged to provide gradients for the optimizable parameters:
\begin{equation}
\nabla_{\theta,f}\mathcal{L}_{\text{SDS}}=\mathbb{E}_{t,\boldsymbol{\epsilon}}\left[w(t)\left(\boldsymbol{\hat{\epsilon}}\left(\boldsymbol{z}_{t} ;t,\boldsymbol{y}\right)-\boldsymbol{\epsilon}\right)\frac{\partial\boldsymbol{z_{t}}}{\partial\theta}\right],
\label{equ:10}
\end{equation}
where $w(t)$ is a weighting function, $\boldsymbol{y}$ is the text prompt, $\boldsymbol{z}_{t}$ is the latent code of the rendered image, with a noise level $t$.
The total loss can be formulated as $\mathcal{L}_\text{total} = \alpha_1\mathcal{L}_{\text{MSE}}+\alpha_2\mathcal{L}_{\text{SDS}}$. By combining implicit representation with SDS optimization, we render high-quality 3D assets within 2D images.

\subsection{Animate Proxy Geometry}
\label{subsec: 3}

The proposed proxy representation is not only amenable to efficient rendering, but also naturally supports standard animation operations. Since the sparse proxy nodes $\mathbf{V}$ are triangulated and rendered via barycentric interpolation as described in Sec.~\ref{subsec: 2}, they form a lightweight mesh-like structure. This enables direct application of both AI-based and artist-authored pipelines for rigging, skinning, and animation on top of our proxy geometry, without requiring a high-quality watertight mesh.

\noindent\textbf{Interactive Animation.} For user-driven animation, we provide an \emph{Interactive Animation} mode that allows direct manipulation of the proxy geometry. The user can specify target positions and trajectories for a subset of proxy nodes (e.g., handle points on limbs or semantic parts), which we treat as positional constraints. To propagate these constraints to the remaining nodes and obtain globally coherent deformations, we employ a position-based dynamics (PBD) scheme. Concretely, we define a set of geometric constraints over the proxy graph, such as edge-length preservation and local rigidity, and iteratively update node positions to satisfy both user-specified constraints and physical priors. This process yields physically plausible deformations of the proxy geometry while preserving the overall structure of the object. Because texture is represented in a distributed manner on the proxy nodes and decoded via the implicit renderer, the resulting animations exhibit high temporal consistency and texture stability under direct user interaction.

\noindent\textbf{Generative Animation.} In addition to explicit user control, we also support a \emph{Generative Animation} mode driven by pretrained motion generation models. We leverage Puppeteer to automatically perform rigging and skinning on our proxy geometry, assigning a skeletal structure and skinning weights to the triangulated proxy nodes. To facilitate efficient reuse of common motions, we construct a motion library containing skeleton templates and corresponding motion sequences for frequently occurring object categories (e.g., humans, quadrupeds, articulated toys). Users can directly select an appropriate skeleton–motion pair from this library, which is then retargeted to the rigged proxy geometry, enabling one-click animation of the target object. Moreover, we integrate AnyTop for prompt-based motion generation. Given a textual description of the desired action, AnyTop produces a motion sequence in skeletal space, which we retarget to the proxy skeleton obtained from Puppeteer. The resulting joint trajectories drive the proxy nodes via skinning, and the final frames are rendered through our proxy-embedding renderer. In this way, our framework supports both library-based and prompt-based generative animation, while maintaining strong identity preservation, geometric coherence, and texture consistency across time. Combined with the lightweight nature of the proxy representation, this allows users to synthesize high-fidelity, controllable 3D animations at low computational cost.

\subsection{Foreground–Background Disentangled Rendering}
After driving the foreground object, we need to re-render and composite it back into the original image. Since our method takes only a single image as input, object motion inevitably reveals previously occluded regions, resulting in missing content (holes) in the background. To address this issue, we adopt a foreground–background decoupled animation compositing strategy.

\noindent\textbf{Background Rendering.} 
For the background, the primary objective is to complete regions that were previously occluded by the foreground. In our hierarchical representation, the proxy nodes associated with the background densely cover the entire image domain. Moreover, the proxy nodes that lie behind the foreground (i.e., within the hole regions) can acquire texture features consistent with those of the visible background nodes via the local latent code, which serves as an information bridge. As a result, the background layer can be obtained simply by rendering the background proxy nodes and their associated texture features over the full image plane, without requiring any additional post-processing.

\noindent\textbf{Foreground Rendering.} 
For the foreground, we project the driven proxy geometry defined in 3D space onto the image plane (cf. Eq.~\ref{equ:6}), and synthesize pixel-space appearance via an implicit rendering mechanism. Specifically, each proxy node carries a distributed feature embedding that locally modulates the rendered texture around its 2D projection, allowing the renderer to recover fine-grained details while remaining robust to imperfections in the coarse 3D estimate. Given a sequence of animated 3D proxy geometries, we apply the same projection-and-rendering procedure frame by frame to produce a temporally coherent foreground stream in the image domain. Finally, for each time step, we composite the rendered foreground with the corresponding background result using a consistent blending strategy (e.g., alpha-based compositing guided by the projected support of proxy nodes), yielding the full animation sequence with both accurate target motion and visually plausible scene completion.

\section{Experiment}
\label{sec:exp}
\subsection{Experimental Setup}
We employ SAM~\cite{kirillov2023segany} to segment the target object and extract its mask. For implicit representation, the texture feature dimension is set to $d=16$, and the positional encoding uses $L=10$. The decoder MLP consists of 8 layers with a hidden dimension of 128 and an output dimension of 3. For SDS optimization, we leverage the SD3 model~\cite{esser2024scaling} for prior, use the Adam optimizer with a learning rate of 0.01, and set the loss weighting factors to $\alpha_1 = 1000$, $\alpha_2 = 0.005$. We optimize the parameters for 1000 iterations on 2 NVIDIA GeForce RTX 3090 GPUs.
\vspace{-0.3cm}
\subsection{Comparison with Existing Methods}
\noindent\textbf{3D Editing.} We present qualitative comparative results for the 3D editing task in Fig.~\ref{fig: quality}. Our method is compared with several state-of-the-art open-source approaches (\emph{i.e.}, 3DIT~\cite{michel2023object}, DragDiff~\cite{shi2024dragdiffusion}, ImgS~\cite{yenphraphai2024image}) with 3D editing capabilities. As shown in Fig.~\ref{fig: quality}, our approach achieves the best performance in both geometric quality and texture consistency. Furthermore, our method supports free-form rotation around arbitrary axes, while 3DIT~\cite{michel2023object} and DragDiff~\cite{shi2024dragdiffusion} fail to support this functionality, revealing clear limitations in their editing flexibility. We employ CLIP-I to evaluate semantic consistency between input and edited images, while SSIM and LPIPS are used to measure structural preservation and perceptual similarity, respectively. Quantitative results (Table~\ref{tab:3d_comparison}) demonstrate that our method outperforms all compared approaches in all metrics.

\noindent\textbf{3D Animation.} To evaluate the animation capabilities, we demonstrate its performance across two primary modalities: rig-driven animation and physical simulation. Unlike existing video generation models that often treat motion as an implicit latent transition, our method explicitly decouples geometric deformation from appearance synthesis.

\begin{figure*}[ht]
    \centering
    \includegraphics[width=1.0\linewidth]{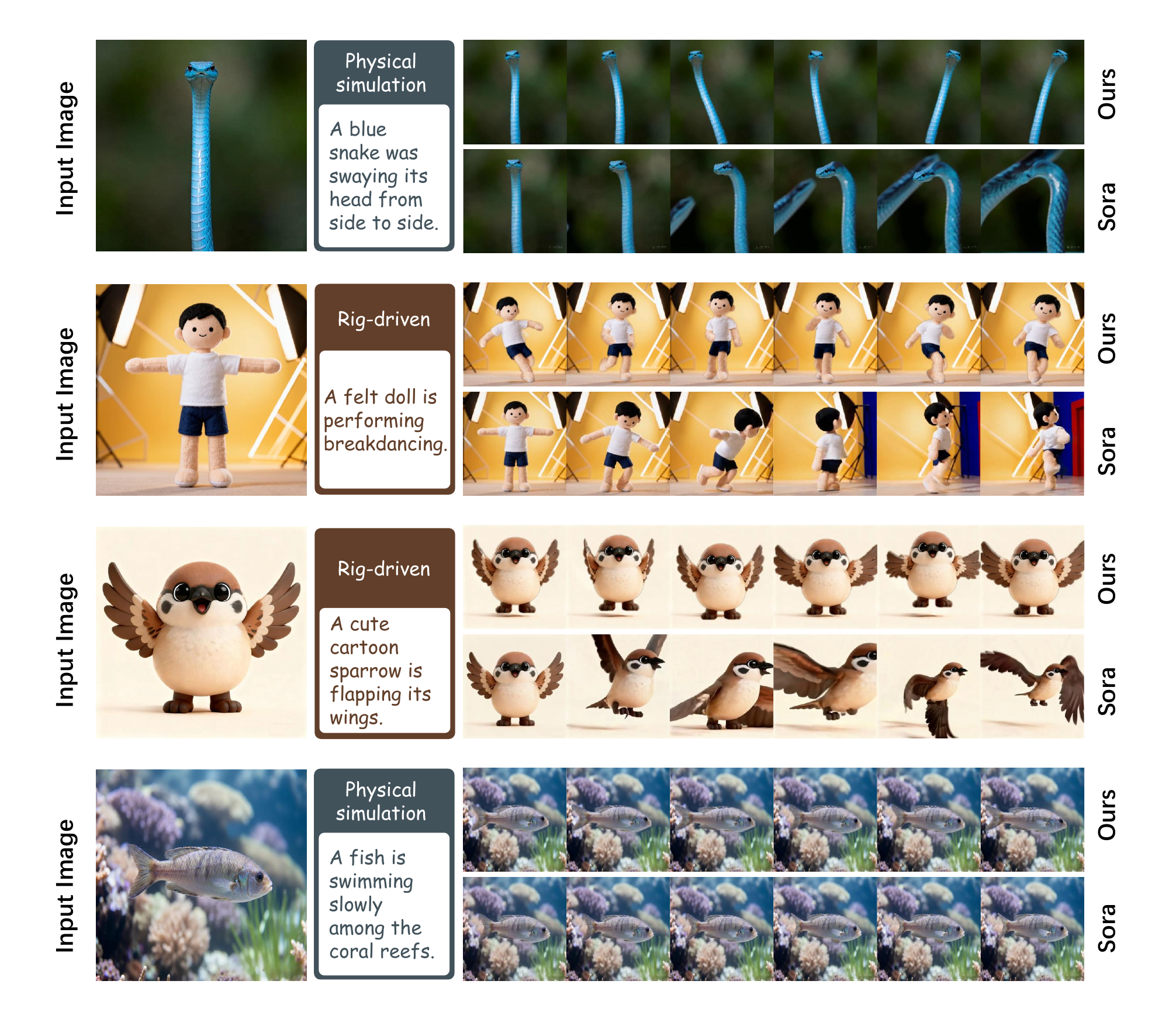}
    % \vspace{-0.4cm}
    \caption{\textbf{Qualitative comparison between Sora and our proposed method across diverse animation tasks.} Sora often struggles with precise motion control and physical plausibility in complex scenarios. In contrast, our framework supports \textbf{rig-driven control} (e.g., the breakdancing felt doll) and \textbf{physical simulation} (e.g., the swimming fish and swaying snake), ensuring superior temporal consistency and structural integrity. Our method demonstrates a higher degree of \textbf{interpretability and controllability} by explicitly modeling the underlying 3D skeletal or physical constraints.}
    \label{fig: comparison}
\end{figure*}

As illustrated in Fig.~\ref{fig: comparison}, for articulated objects such as the \textit{felt doll} and the \textit{sparrow}, our model accurately follows the input skeletal sequences. The 2D-3D aligned proxy representation ensures that the character's structural integrity is preserved even during complex, large-scale movements (e.g., breakdancing). Unlike video generation models such as Sora, which frequently exhibit structural distortions or unrealistic morphing artifacts when handling out-of-distribution motions, our rig-driven approach ensures rigorous geometric integrity. By providing deterministic control over every joint, our framework bypasses the stochastic nature of end-to-end generation, making it more robust and suitable for professional production-level character animation.

% While Sora can generate visually pleasing videos, it frequently exhibits structural distortions or "morphing" artifacts when the motion deviates from the training distribution. Our rig-driven approach, however, provides deterministic control over every joint, making it more suitable for production-level character animation.

Furthermore, in scenarios involving physical priors, such as the \textit{swimming fish} and \textit{swaying snake}, our framework successfully integrates physical simulation into the animation pipeline.

By manipulating the 3D proxy nodes according to physical kinematic equations (e.g., forward kinematics for articulated motion or mass-spring systems for deformable structures), we generate animations that exhibit superior structural consistency and physical plausibility. Unlike the stochastic motion synthesis in Sora, which is prone to \textbf{geometric collapse} and \textbf{unrealistic morphing}, our approach ensures that the dynamic trajectories are governed by explicit physical laws. This deterministic framework allows for the synthesis of complex behaviors—such as the swaying of a snake or the rhythmic swimming of a fish—while maintaining strict \textbf{temporal stability} and \textbf{identity preservation} of the target object.

% By manipulating the 3D proxy nodes according to fluid dynamics or wave functions, we generate animations that are not only visually consistent but also physically plausible. Notably, our method maintains high-fidelity texture mapping throughout the animation; the fine details of the fish scales and snake skin remain temporally stable without the "texture sliding" or flickering issues common in pure generative methods. These results highlight that by anchoring image-space priors to a 3D structural carrier, 3DProxyImg achieves a superior balance between creative generation and rigorous 3D control.

% \begin{table}[t]
% \centering
% \begin{tabular}{lccccc}
% \toprule
% Method & PSNR & SSIM & LPIPS & FID & CLIP \\
% \midrule
% OURS & 18.363 & \textbf{0.777} & \textbf{0.192} & 76.220 & 0.961 \\
% 3DIT & \textbf{20.179} & 0.658 & 0.245 & 89.728 & 0.923 \\
% DRAGDIFF & 19.741 & 0.717 & 0.208 & \textbf{58.525} & \textbf{0.973} \\
% IMAGE_S & 17.525 & 0.659 & 0.277 & 161.856 & 0.901 \\
% \bottomrule
% \end{tabular}
% \caption{Quantitative comparisons on our dataset.}
% \label{tab:quant}
% \end{table}
\vspace{-0.3cm}
\begin{table}[ht]
  \centering
  \caption{Quantitative comparison with other methods.}
  \label{tab:3d_comparison}
  \small
  \begin{tabular}{lcccc}
    \toprule
    \textbf{Metric}& \textbf{3DIT} & \textbf{DragDiff} & \textbf{ImgS} & \textbf{Ours} \\
    \midrule
    SSIM $\uparrow$ & 0.658              & 0.717            & 0.659                   & \textbf{0.777} \\
    LPIPS $\downarrow$               & 0.245              & 0.208           & 0.277                  & \textbf{0.192} \\
    CLIP-I $\uparrow$       & 0.886              & 0.955            & 0.897                   & \textbf{0.961} \\
    % Mean Confidence $\uparrow$        & 0.638              & 0.669            & 0.611                   & \textbf{0.740} \\
    \bottomrule
  \end{tabular}
\vspace{-0.3cm}
\end{table}

\subsection{Ablation Study}
\begin{figure}[ht]
    \centering
    \includegraphics[width=1.0\linewidth]{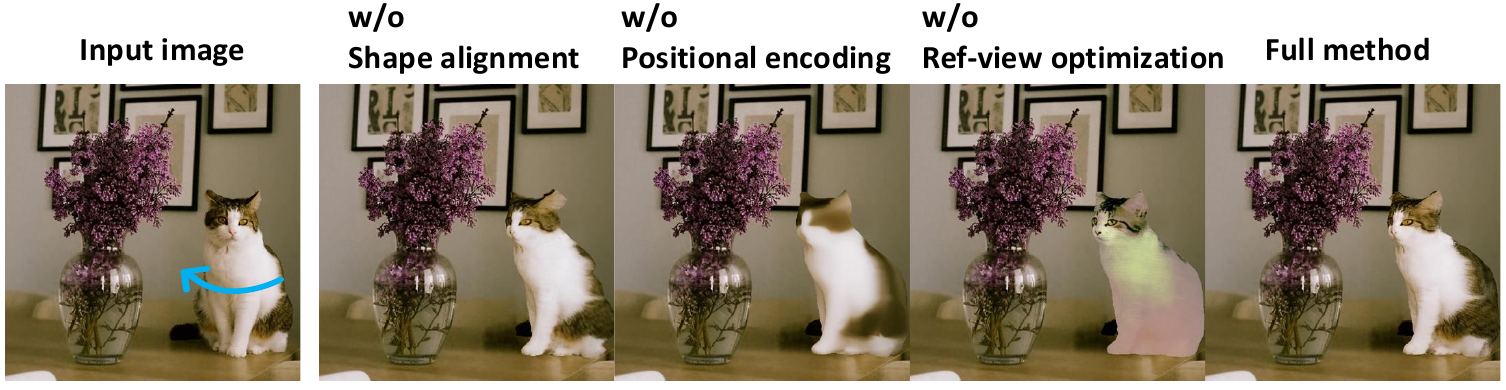}
    \vspace{-0.4cm}
    \caption{Ablation study on the effect of our framework.}
    \label{fig: ablation}
\end{figure}

The following experiments are designed to validate the efficacy of our framework. As shown in Fig.~\ref{fig: ablation}, removing the shape alignment algorithm leads to noticeable geometric distortions in the object shape; omitting positional encoding during rendering results in the loss of high-frequency details; and excluding the reference viewpoint during optimization (MSE loss) causes a degradation in texture consistency of the target object. Integrating these key components enables our method to achieve both accurate geometric alignment and high-fidelity texture appearance.
\vspace{-0.3cm}
\section{Conclusion}
\label{sec:conclusion}
In this work, we revisited single-image 3D animation generation through the lens of the inherent trade-off between rendering quality and 3D controllability, and proposed a lightweight framework that sidesteps this limitation. By introducing a 2D–3D aligned proxy representation, our method uses only a coarse 3D estimate as a structural carrier while delegating high-fidelity appearance and view synthesis to powerful image-space generative priors. This decoupled design enables 3D-aware motion control, interactive editing, and coherent background animation that are comparable to traditional 3D pipelines, yet without requiring accurate geometry or expensive optimization. Extensive experiments show that our approach supports efficient animation generation even on low-power platforms, and consistently outperforms video-based 3D animation methods in identity preservation, geometric and textural consistency, and the level of precise, user-controllable interaction it affords. We believe this proxy-based paradigm offers a promising direction for scalable, accessible 3D animation tools and can be further extended to more complex scenes, multi-object interactions, and broader content creation workflows.

\bibliographystyle{plain}
\bibliography{sn-bibliography}% common bib file
% \input sn-article.bbl

\end{document}